# A Legged Soft Robot Platform for Dynamic Locomotion

Boxi Xia[1, †], Jiaming Fu[1, †], Hongbo Zhu[1, †], Zhicheng Song[1, †], Yibo Jiang[1], Hod Lipson[1]

*Abstract*— This paper presents an open-source untethered quadrupedal soft robot platform for dynamic locomotion (e.g., high-speed running and backflipping). The robot is mostly soft (80 vol.%) while driven by four geared servo motors. The robot's soft body and soft legs were 3D printed with gyroid infill using a flexible material, enabling it to conform to the environment and passively stabilize during locomotion in multi-terrain environments. In addition, we simulated the robot in a real-time soft body simulation. With tuned gaits in simulation, the real robot can locomote at a speed of 0.9 m/s (2.5 body length/second), substantially faster than most untethered legged soft robots published to date. We hope this platform, along with its verified simulator, can catalyze agile soft robots' development.

## I. Introduction

Legged locomotion is perhaps the most prevalent mode of locomotion for land animals. The locomotion of legged animals benefits from their compliant structures, which passively deform to absorb impact and adapt to different terrains. Research on dynamic locomotion for legged robots has been primarily focused on rigid mechanisms. Many motor-driven legged rigid robots capable of highly dynamic locomotion [1]–[5] have been developed. For example, the Mini Cheetah quadrupedal robot, driven by low-transmission-ratio geared servo motors, can run at high speed and backflip using model predictive control [5], [6]. The ANYmal quadrupedal robot is actuated by series elastic actuators, and it can robustly locomote over challenging terrains [3], [7].

Rigid robots can be programmed with active compliance via accurate sensors and appropriate feedback control to adapt to the environment [8]. In comparison, soft robots can take advantage of their deformable structure to achieve passive compliance in addition to programmed active compliance, making the soft robots potentially more adaptable to different terrains.

Unfortunately, research on dynamic (airborne) locomotion for soft robots is limited. As listed in TABLE I, there are several high-speed untethered legged robots made with flexible frames, there are only a few soft robots capable of fast locomotion with speed greater than one body length per second (BL/s). The existing fast soft robots are generally insect-scale and tethered, having little space for onboard sensing.

*Research supported by the U.S. Defense Advanced Research Project Agency (DARPA) TRADES grant number HR0011-17-2-0014, and Israel Ministry of Defense (IMOD) grant number 4440729085 for Soft Robotics.

[1] Department of Mechanical Engineering, *Columbia University*, New York, NY. Corresponds to *bx2150@columbia.edu* (B.X)

[†] These authors contributed equally to this work.

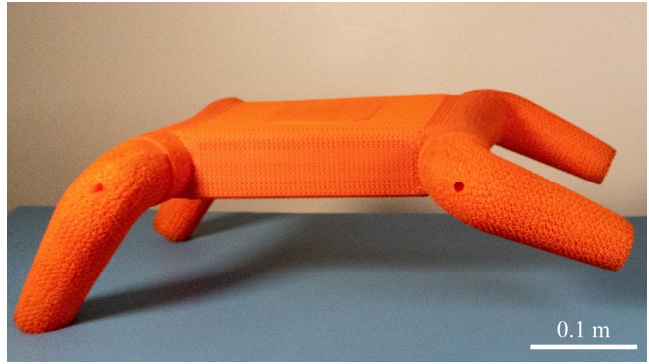

Fig. 1. The Flexipod running with bounding gait

There are several challenges for developing fast soft robotic runners: (1) developing reliable, high-speed, and high-power density actuators for soft robots; (2) developing realistic and fast simulations for soft matter dynamics; (3) developing rapid and reliable free-form manufacturing of soft robots.

Many soft actuation methods have been developed. Soft pneumatic and hydraulic actuators exhibit high actuation force and moderately high speed while requiring external pumps to actuate [9], [10]. By comparison, shape memory alloy (SMA) can be directly actuated through resistive heating. While SMA offers higher power density, it is limited by slow speed and low efficiency [11], [12]. The stacked dielectric elastomer actuator (DEA) is a promising technology as it exhibits fast actuation speed, high strain, and high power density [13]–[15]. Stacked DEA consists of alternatingly stacked thin (<100 microns) dielectric elastomer layers and conductive electrodes, and high voltage is applied on the electrodes [13], Currently, stacked DEA is challenging to manufacture and less robust to damage. Other soft actuation methods such as thermally driven[16] and light actuated [17] have limited speed or power density and may not be suitable for highly dynamic soft robots.

On the one hand, soft robots may be robust to impact but are limited by the lack of accessibility to fast and powerful soft actuators (e.g., stack DEA); on the other hand, conventional rigid actuators such as brushless DC motors (BLDC) are fast, dense in power and widely available, while being less robust to impact. Combining motors with soft structural material may help create fast soft robots with good manufacturability. In fact, TABLE I lists several high-speed flexible robots that are actuated with motors [18]–[21].

Due to large deformation, controlling a high-speed soft robot can be challenging. A robust soft robot controller may require explicitly or implicitly modeling its soft body dynamics. Empirically, a soft robot's dynamic model can be obtained via experimenting on the actual robot; however, it can be resource and time expensive. In this case, physics-based soft body simulation provides an attractive alternative. PyBullet [22] and MuJoCo [23] have been used extensively in research for rigid body simulation; however, they have limited support for volumetric soft body simulation. Compared to rigid body simulation, soft body simulation has a much higher number of degrees of freedom and thus is more computationally expensive. Several methods are suitable for large scale real-time soft body simulation: The Titan Simulator [24] is a CUDA accelerated spring-mass-based soft body simulation engine. The Nvidia Flex [25] is a simulation engine using position-based dynamics with unified particle representation. The Taichi [26] is based on the moving least squares material point method, which offers differentiable physics suitable for machine learning. Compared to the Titan simulator, Nvidia Flex can simulate more types (liquid and rigid) of objects, and Taichi's simulation is differentiable, while the Titan simulator offers faster simulation speed.

TABLE I. SOME SOFT/FLEXIBLE ROBOTS

| Robot | Year | Soft? | Tethered? | Actuation | Actuator count | BL (m) | Max Speed (BL/s) | Max Speed (m/s) | Image |
|---|---|---|---|---|---|---|---|---|---|
| Flexipod (this work) | 2021 | 80 vol.% soft | Untethered | motor | 4 | 0.36 | 2.5 | 0.9 | 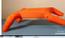 |
| Resilient quadruped [41] | 2014 | Mostly soft | Untethered | pneumatic | 6 | 0.65 | 0.008 | 0.005 | 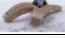 |
| multi-gait caterpillar [42] | 2018 | soft | Untethered | SMA | 6 | 0.074 | 1.0 | 0.074 | 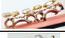 |
| DEAnsect [14] | 2019 | soft | Untethered | DEA | 3 | 0.04 | 0.75 | 0.03 | 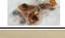 |
| Inchworm [43] | 2017 | soft | tethered | DEA | 1 | 0.02 | 1.0 | 0.02 | 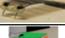 |
| Hopping-runner [44] | 2019 | mostly soft | tethered | DEA | 2 | 0.085 | 6.1 | 0.52 | 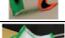 |
| bistable crawler [9] | 2020 | soft | tethered | pneumatic | 2 | 0.07 | 2.7 | 0.19 | 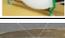 |
| Robust insect [45] | 2019 | flexible | tethered | piezoelectric | 1 | 0.01 | 20 | 0.2 | 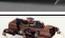 |
| X2-VelociRoACH hexapod [19] | 2015 | flexible | Untethered | motor | 1 | 0.104 | 47 | 4.9 | 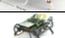 |
| HAMR-F quadruped [46] | 2018 | flexible | Untethered | piezoelectric | 8 | 0.045 | 3.8 | 0.172 | 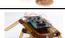 |
| RoACH hexapod [47] | 2008 | flexible | Untethered | SMA | 2 | 0.03 | 1.0 | 0.03 | 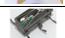 |
| DASH hexapod [18] | 2009 | flexible | Untethered | motor | 2 | 0.1 | 15 | 1.5 | 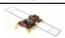 |
| VelociRoACH hexapod [20] | 2013 | flexible | Untethered | motor | 2 | 0.1 | 27 | 2.7 | 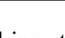 |

Finally, free-form manufacturing of soft robots has been challenging. Compared to subtractive manufacturing, 3D printing enables rapid prototyping for objects with complex structures and functionalities. 3D printing enables designs to be open-sourced and fabricated with little specialized tooling. For example, robotic platforms such as the "Poppy" humanoid robot [27] and the "Open Dynamic Robot Initiative" quadruped [28] used 3D printed rigid parts extensively. Advances in soft and flexible 3D printing have opened new possibilities for fabricating soft robot directly. Many 3D printed soft actuators, sensors, and control circuits have been proposed [29]–[32]. While some of the technologies are still nascent, materials such as Thermoplastic polyurethanes (TPU) are widely available and can be printed with an off-the-shelf fused deposition modeling (FDM) 3D printer. By varying the infill density, an FDM printer can fabricate lattices with varying meta-material stiffness.

In this paper, we present an open-source soft quadrupedal robot platform (named Flexipod) for researching soft dynamic locomotion. As shown in Fig. 1, this robot combines the merit of both soft material and rigid BLDC actuators. The Flexipod can be easily manufactured using a desktop 3D printer and off-the-shelf flexible TPU material. The soft body and legs are compliant and can absorb impact during locomotion, thereby protecting the motors and electronics. Driven by BLDC servo motors, the Flexipod can dynamically locomote at 0.9 m/s (2.5 BL/s) and perform backflips. Locomotion in various terrains was tested, as shown in the following section. Besides, we developed and verified a real-time soft body simulation based on Titan [24], which enabled rapid design improvements and gait tuning.

This paper's main technical contributions are: (1) an untethered motor-actuated soft legged robotic platform for agile soft-body locomotion. The robot's soft components were 3D printed with flexible filament using gyroid infill patterns, which is easier to manufacture compared to molding. The printed components' softness can be reduced by adjusting the 3D printing flow rate and infill density. (2) A real-time soft body simulation software for simulating soft robots with revolute joints. While this paper demonstrated simulating a soft robot with four revolute joints, the simulation can be extended to robots with more joints.

The hardware design, code, and videos are available at (https://boxixia.github.io/Flexipod).

## II. MECHANICAL DESIGN

### A. Material

The structural components of the Flexipod were 3D-printed using a desktop 3D printer (Troodon Core-XY printer, VIVEDINO). All components were printed with a standard 0.4 mm nozzle at 0.2 mm layer height.

The flexible material used for printing the soft body and legs is a flexible TPU (Cheetah flexible filament, Ninjatek). This filament is highly stretchable (elongation: at yield = 55%, at break = 580%) and can undergo frequent deformation.

Unlike many soft FDM filaments that require slow speed or specialized print-head, this filament can be printed with a normal print setting.

Although the filament has a shore hardness of 95A, it is possible to achieve a lower hardness by varying the infill density and the flow rate (the amount of the material extruded is multiplied by this factor). Gyroid infill pattern [33] was used because it has near isotropic properties and relatively high strength [34]. Fig. 2 (b-c) shows the cross-section of two cylinders printed with gyroid infill. It qualitatively illustrates that the infill's thickness can be adjusted by altering infill density and flow rate, affecting the overall mechanical properties such as the metamaterial's hardness and stiffness.

Fig. 2 (a) shows the influence of flow rate and infill density on 3D-printed cylinders' shore hardness with gyroid infill and a single skin layer. The shore hardness was measured using a Shore A durometer (Durometer Tester Hardness Tester 10-90HA, Hilitand). Specifically, 16% and 20% infill density were examined. Since the tester can only measure a printed soft sample with skin, this result provides an upper bound to estimate the shore hardness of infill-only (without skin) 3D printed soft parts. It shows a promising possibility for direct printing soft parts using semi-soft filament by an off-the-shelf FDM printer.

For a flexible print, skins (densely spaced surface layers) can significantly reduce flexibility. Thus, to achieve high flexibility, skin layers were only printed on surfaces that require attachment with other parts. All parts were printed using the gyroid infill. The Flexipod soft body was printed with 20% infill density and 90% flow, while the softer leg was printed with 16% infill density and 80% flow.

*B. Body design*

The Flexipod's total mass is 3.21 kg, and its body is 0.360 m long and 0.214 m wide. 96 vol. % of the 3D printed parts of the robot are soft, making the robot is 80 vol.% soft. The Flexipod is made of three parts: a soft body with motor modules inside, soft legs, and electronics. Fig. 3 shows a partially assembled Flexipod exhibiting its internal components.

The soft body can be assembled from several 3D printed parts: front and back covers, lower and upper mid-body, and top cover. The soft body's 10 mm gyroid-patterned wall can passively deform to dampen shock, thereby protecting the electronics inside the main body and enhancing the robot's robustness. The soft cover on the top of the body can be opened for installing the electronics and recharging the battery. A front-facing camera is embedded for video streaming.

*C. Leg design*

The soft legs design's primary consideration was to achieve high step length, low mass, low inertia, and appropriate flexibility. The leg has a step length of 158 mm, a minimum 48 mm diameter at the tip, and a maximum 64 mm diameter at the shoulder. We compared the 3D printed leg ($0.2 g/cm^3$) with a leg molded with foamed-silicone (Smooth-on Equinox™ 38 MEDIUM mixed with foam beads, $0.7 g/cm^3$) of the same geometry, the direct 3D printed leg (Fig. 4 (a)) is 71 % lighter. The leg's hollowed structure (Fig. 4 (b)) reduces its moment of inertia and mass (80 g), which also lowers the motor load. Compared to a non-hollowed leg, hollowed leg deforms more easily. It also acts as a passive damper. Fig. 4 (c) demonstrates the damping effect, where we drop the Flexipod from 2.0 m height. At impact, the soft legs were compressed, and the body was bent inward to absorb the impact. Thus, the Flexipod can restore to a normal posture from the fall on its own.

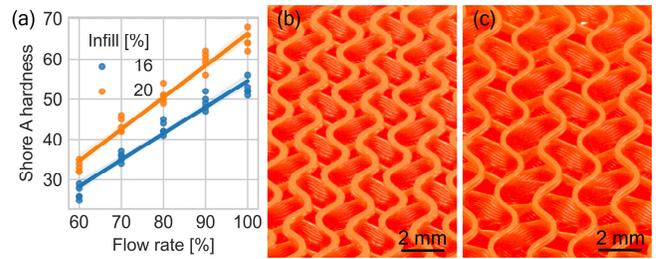

Fig. 2. (a) Shore hardness vs. flow rate and infill density; (b-c) cross-section of 3D printed cylinders with (b) gyroid 20% infill density and 90% flow rate, and (c) with 16% infill density and 80% flow rate.

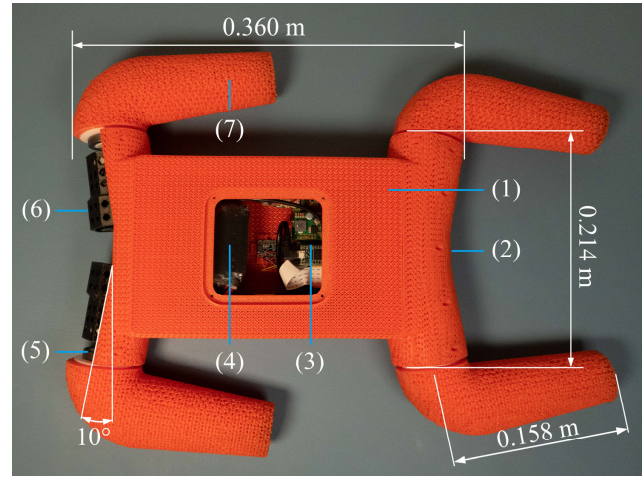

Fig. 3. Structure of the soft robot: (1) Soft main body; (2) Camera; (3) Electronic components; (4) Li-Po battery; (5) Bearing; (6) DJI M3508 brushless DC motor enclosed in a motor shell; (7) Soft leg.

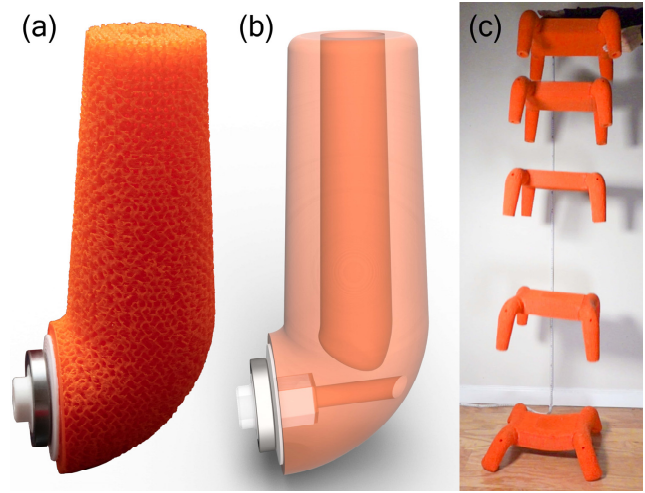

Fig. 4. Flexipod soft leg and demonstration: (a) Soft Leg, leg coupler and bearing; (b) 3D model of the leg assembly showing its internal structure; (c) chronophotograph of the Flexipod dropped from 2.0 m.

Each leg is actuated by a BLDC servo motor (M3508, DJI). The motor weighs 386 g and provides $3\ N \cdot m$ torque at 469 revolutions per minute (rpm). It was enclosed in a motor shell, and its shaft was connected to a soft leg by a leg coupler. Each leg coupler was glued to a leg with super glue (102433 XL Super Glue Gel, Gorilla). The sealed ball bearings (6806-2RS,30x42x7mm) were installed between the couplers and the soft body to offload the radial load. The motor shells and leg couplers were printed with rigid filament (PolyMax™ PC, Playmaker).

## III. ELECTRICAL DESIGN

As shown in Fig. 5, the electrical system enables wireless control of the Flexipod using a computer. Control commands and robot state feedback are serialized into binary MessagePack [35] packets and transported via UDP. Onboard video is captured from a camera (Raspberry Pi NoIR Camera Module v2) and streamed via WIFI to a PC. The Raspberry Pi acts as a WIFI access point to directly communicate with the computer to reduce latency.

The motion control and robot state estimation were implemented on a Teensy 4.0 board, which communicates with the Raspberry Pi via USB serial. The Teensy receives position command from the Raspberry Pi and then controls 4 ESCs (electric speed controller) via CAN (Controller Area Network) bus interfaced through a CAN bus transceiver (MCP2551, Comimark). The ESCs output three-phase power in specific frequencies to drive the motors. The motors' feedback includes angular position and velocity, current, temperature over the CAN bus to the Teensy. Motors' data, along with the IMU measurements, are transported to PC via UDP.

An IMU (LSM6DSOX + LIS3MDL, Adafruit) is attached at the center of the body, and it measures acceleration, angular velocity, and magnetic field and communicates with the Teensy via I2C. The robot orientation is estimated using the Madgwick filter [36].

A LiPo battery (Ovonic 1550 mAh 22.2V 6S 100C, 274g) powers the ESCs directly; it also powers other electronics through a 5V step-down voltage regulator (D36V50F5, Pololu). A custom PCB connects Teensy, Raspberry Pi, voltage sensor, CAN transceiver, voltage regulator, and the battery. To avoid the battery be over-discharged, the Teensy 4.0 monitors the battery voltage using a voltage sensor (INA260, Adafruit) through an I2C bus.

## IV. CONTROL

As shown in Fig. 6, the Flexipod implements a classical joint-space position control. For each motor, the desired motor shaft position $\theta_{des}$ is continuously sent from the PC to the robot to track the position command. $e_\theta$ is the error between command and the measured motor shaft position. A proportional controller receives this error, then output $d$ to be subtracted by the speed feedback $\omega_{meas}$. Then the angular velocity error $e_\omega$ is received by a PI controller. The resulted current $i_{des}$ after this step is sent to the electric speed controller for controlling the motor directly. After the motor outputs the rotation angle $\theta_{real}$, the encoder sends motor angular position $\theta_{mes}$ and velocity $\omega_{meas}$ to the beginning of the control loops.

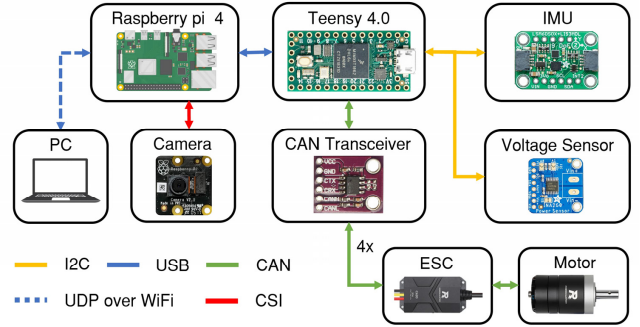

Fig. 5. The electrical system of the Flexipod.

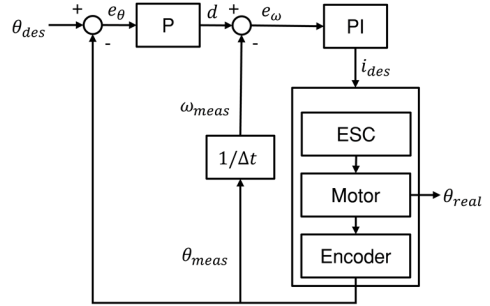

Fig. 6. Schematics of the Flexipod controller. (Subscripts: des-desired, meas-measured, real-real value)

## V. SIMULATION

### A. Exploratory study

Initially, we used rigid body simulation for a coarse study of the leg angle's kinematic effect on robot locomotion performance. Although rigid body simulation cannot model the effect of deformation, it computes faster than soft body simulation, and it produces approximate results at low speed and momentum, appropriate for initial evaluation.

The simulation was conducted in Pybullet [22] to examine the leg angle's influence on the robot velocity during forward locomotion, and its body's angular velocity during rotation. The leg angle $\alpha$ is the angle between the leg joint axis and the body's median plane, as shown in Fig. 7 (a). A low-speed locomotion test was conducted by simulating the robot with $\alpha$ over 0° to 30° with 2° increment while keeping the same mid-body and leg configuration. The average joint velocities were set at $2\pi\ rad/s$.

Fig. 7 (b) shows the effect of leg angle $\alpha$ on the robot's velocities in bounding gait, pace gait, and rotation. We found that a smaller leg angle is preferred for fast forward locomotion while a larger leg angle is desirable for rotation. Also, a larger leg angle makes the robot wider, which may be undesirable when traveling in tight spaces. Thus, we decide to build the robot with $\alpha = 10°$ to balance among fast forward locomotion, fast turning, and compact size.

### B. Soft-body simulation

For a higher fidelity simulation, deformation must be taken into account. We developed a real time soft robot simulation environment based on the Titan [24], which is a CUDA [37] accelerated massively parallel asynchronous spring-mass simulation library. We extended the Titan library with a rotational kernel for contact-coupling of soft bodies. The

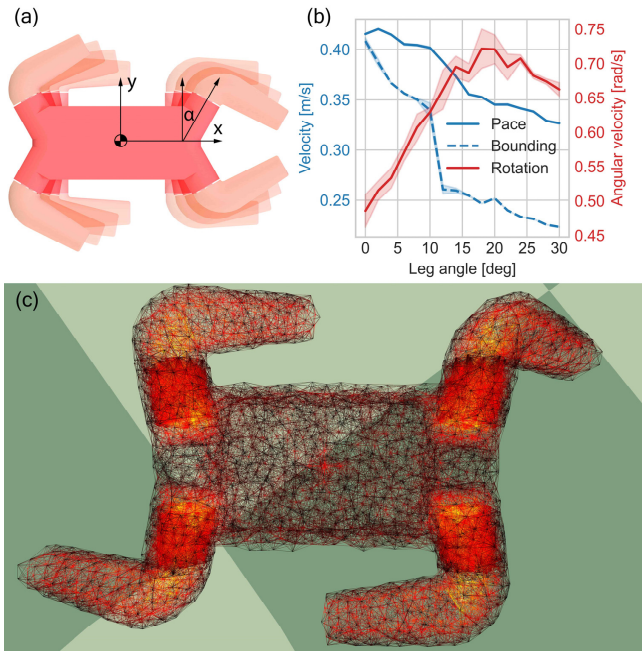

Fig. 7. Flexipod simulation setup and result: (a) Flexipod with different leg angle $\alpha$ overlayed; (b) Locomotion benchmark for robots with varying leg angle $\alpha$: pace gait, bounding gait, and rotation; (c) Flexipod soft body simulation

Flexipod simulation achieved a speed of $1.4 \times 10^9$ spring evaluations per second on a consumer Nvidia 2080Ti GPU. The soft body simulation for the Flexipod is shown in Fig. 7 (c). The simulated Flexipod consisted of 4.7k masses and 70k springs, and it was simulated in real time at a physics update time step of $5 \times 10^{-5}$ second.

The workflow for converting a parametric robot design to a spring-mass representation is as follows: First, point clouds were sampled volumetrically from the body's mesh and legs using Poisson disk sampling [38]. Second, neighboring sampled vertices (masses) were connected by edges (springs), which created each body part's spring-mass representation.

To create rotational joints among the body parts: First, sampling the vertices of two parts of a joint (e.g., body and a leg) within a volume anchored at the rotation axis of this joint, and connecting them with two anchors of the rotation axis. Next, the chosen vertices of one part (e.g., body) were connected randomly with the selected vertices of its corresponding part (e.g., a leg) by springs whose rest lengths were reset every rotation update, acting as a proxy for joint friction. At each joint update, the two parts' vertices were rotated in the opposite direction.

To validate the simulation, we tuned simulation parameters (such as friction and contact stiffness) and compared the robot's average velocities using pace and bounding gait in both simulation and reality. The measurement in reality is done in a sand field. Fig. 8 summarized the average velocities measured within a 3-meter distance. The result shows that the simulation simulates the bounding gait accurately (within 12% error), while there remains some gap for pace gait at low velocity (within 16% error). The simulation error is calculated as the absolute percentage error between the velocity in simulation and reality.

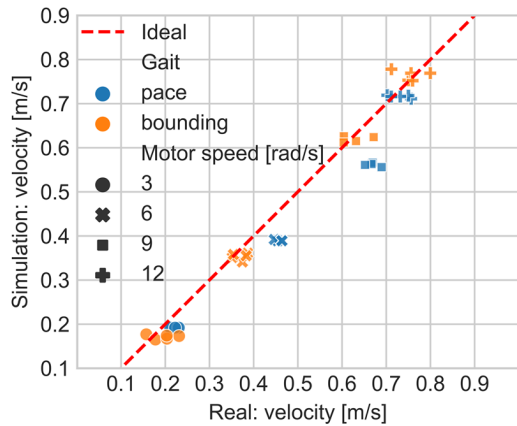

Fig. 8. simulation versus reality: Average velocities were measured within a 3-meter distance in simulation (y-axis) and in a sand-field (x-axis). If the simulation matches reality perfectly, all sampled points should be on the ideal line (dashed red line)

The soft simulation enables rapid evaluation of the robot designs and controllers. Through simulation, we discovered that the Flexipod locomotes faster when using hollowed legs with a large tip as compared to non-hollowed legs with a smaller tip, possibly because hollowed legs are more flexible and have a larger contact area. Robots with different combinations of inner and outer cross-section diameters of the legs were simulated. By comparing robots' forward locomotion speed in simulation, a set of parameters that performed well were chosen. We also tuned gait parameters (described in section VI.A) in the simulation to achieve fast locomotion.

## VI. LOCOMOTION

### A. Gaits

The gaits are generated by a template, where we section the motor position in multiple phases and assign a constant velocity for each phase. A basic trajectory consists of two phases: stance and swing. The stance phase (when a leg is assumed to be on the ground) and swing phase (when a leg is assumed to be in the air) trajectories of the Flexipod are parameterized based on the current position of a leg:

$$\theta_t = \begin{cases} (\theta_L + \omega_s t_c) \bmod 2\pi, & \text{if } \theta_L < \theta_t < \theta_H \\ [\theta_H + \omega_a(t_c - s)] \bmod 2\pi, & \text{otherwise} \end{cases} \quad (1)$$

$$t_c = \left(\hat{t}_c + \frac{\omega dt}{2\pi}\right) \bmod 1 \quad (2)$$

Where $\theta_L$ and $\theta_H$ are the positions when the stance phase starts and ends respectively, $c = \theta_H - \theta_L$ is the contact angle which legs will move through during the stance phase, $t_c$ is the normalized time and $\hat{t}_c$ is the normalized time at the previous time-step, $\omega = 2\pi/t_c$, $\omega_s = c/s$ and $\omega_a = (2\pi - c)/(1 - s)$ are the normalized average speed in one cycle, in the stance phase and the swing phase, respectively. The stance ratio $s$ is the time ratio of the stance phase.

Each leg can be actuated by following an individual trajectory parameterized by the gait template. By arranging appropriate phase offsets and other parameters of the gait template, various gaits [39] can be obtained. Fig. 9 shows the robot with bounding gait, pace gait, turning, and backflipping. Video of the locomotion is available at [40].

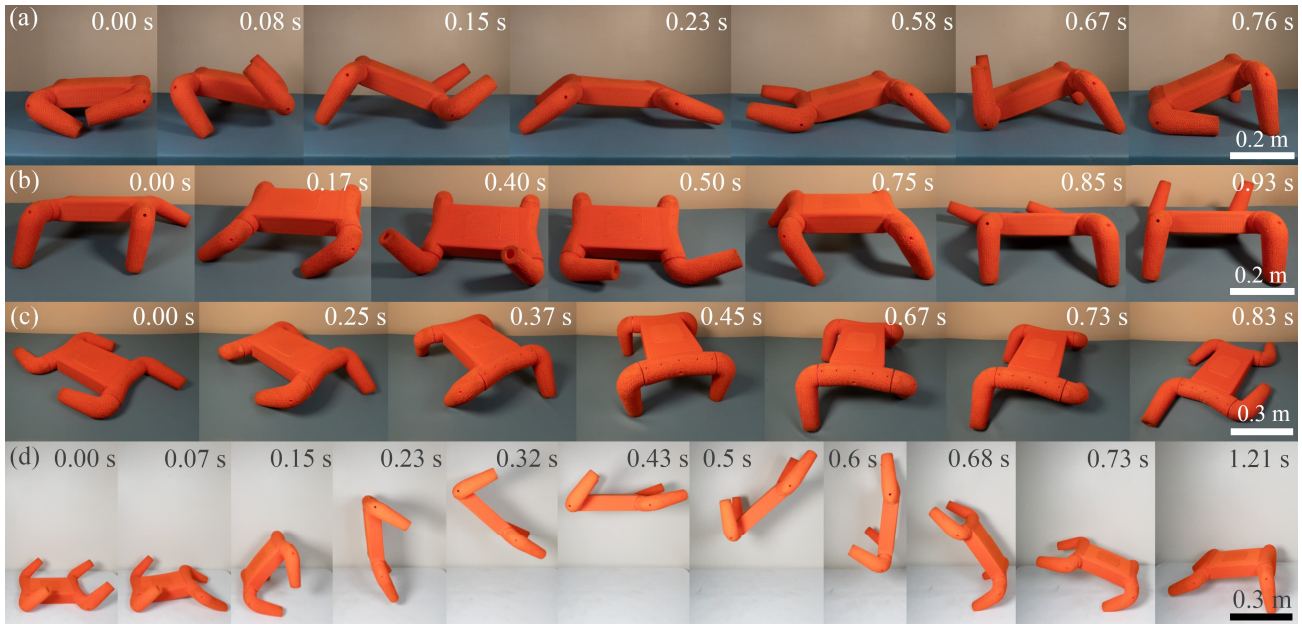

Fig. 9. Flexipod locomotion patterns: (a) bounding gait, (b) pace gait, (c) turning, (d) backflip.

In pace gait, the two legs on the left side of the robot are synced with the same phase, while the right legs are synced with another phase. In one gait cycle, each group of legs' tips contact the ground alternatingly. In bounding gait, the front two legs are synced with the same phase, while the back legs are synced with another phase. When one pair of legs' tips leave the ground, the other pair are already in contact with the ground so that the tips of the front and back legs touch the ground in turns. For turning, all legs are first synchronized in the same phase, and then rotate in the same direction, i.e., clockwise or counter-clockwise, then the robot body can turn left or right according to the rotation direction of legs. The Flexipod is also able to restore to a normal posture from flipping. When the robot is flipped, the robot can adjust its gait according to the estimated orientation and continue its locomotion. Finally, the Flexipod can perform backflips by rotating the front and back legs at high speed and with a timed offset. Fig. 9 (d) shows clear deformation of the legs, which dampens shock at touchdown.

The Flexipod has been tested on multiple terrains with various gaits at different speeds. We observed that the gait in pace is more suitable for flat terrains, such as carpets and concrete floors. Fig. 10 (g) shows the chronograph of the Flexipod running on a carpet using the pace gait. Its velocity reached 0.916 m/s (2.5 BL/s) at a joint velocity of 18 rad/s.

*B. Terrains*

The soft body and soft legs provide protection and damping for the motors and electronics, enabling the Flexipod to explore different terrains robustly. As shown in Fig. 10, the Flexipod can locomote dynamically on various natural and human-made terrains, such as carpet, brick floor, marble floor, grass, and sand.

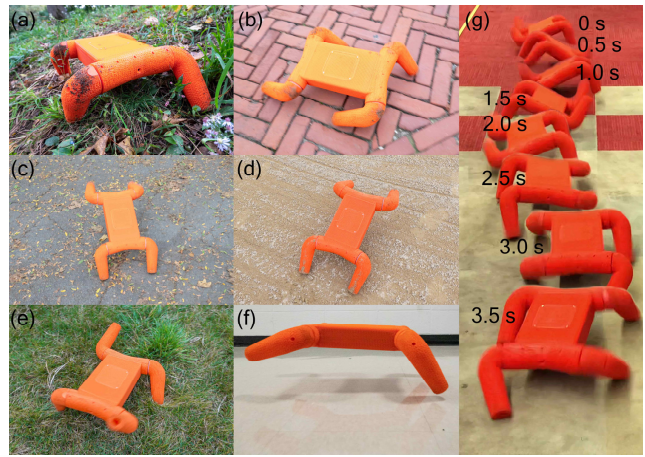

Fig. 10. Flexipod locomotion in (a) natural terrain, (b) brick floor, (c) concrete floor, (d) sand, (e) grass, (f) marble floor, (g) carpet. (chronophotograph)

## VII. CONCLUSION

This paper presented Flexipod, an untethered soft quadrupedal robot driven by motors. Our soft-body design, combined with motors, enables the robot to locomote faster than most legged soft robots in various terrains while having its electronics protected. A real-time soft body simulator was also presented for testing the novel mechanical design and control. We hope our established hardware and software can be used to catalyze the development of agile soft robots.

ACKNOWLEDGMENT

This work is supported by the U.S. Defense Advanced Research Project Agency (DARPA) TRADES grant number HR0011-17-2-0014, and the Israel Ministry of Defense (IMOD) grant number 4440729085 for Soft Robotics.